\documentclass[hidelinks,12pt]{article}
\usepackage{amsmath}
\usepackage{graphicx,psfrag,epsf}
\usepackage{enumerate}
\usepackage[square,numbers]{natbib}
\usepackage{url} 
\usepackage[document]{ragged2e}
\usepackage[backref]{hyperref}

\usepackage[table,xcdraw]{xcolor}
\newcommand{\blind}{0}

\usepackage{amssymb}
\usepackage{bm}
\usepackage{amsthm}
\usepackage[titletoc,title]{appendix}
\usepackage{rotating}
\usepackage[skip=0.5\baselineskip]{caption}
\usepackage{tabularx}
\usepackage{longtable}
\newcolumntype{b}{X}
\newcolumntype{s}{>{\hsize=.5\hsize}X}

\newcommand{\mbb}[1]{\mathbb{#1}}
\newcommand{\mbf}[1]{\boldsymbol{#1}}
\newcommand{\mcal}[1]{\mathcal{#1}}
\newcommand{\mrm}[1]{\textrm{#1}}

\theoremstyle{definition}

\makeatletter
\newcommand\incircbin
{%
  \mathpalette\@incircbin
}
\newcommand\@incircbin[2]
{%
  \mathbin%
  {%
    \ooalign{\hidewidth$#1#2$\hidewidth\crcr$#1\bigcirc$}%
  }%
}
\newcommand{\owedge}{\incircbin{\wedge}}
\makeatother

\begin{document}

\def\spacingset#1{\renewcommand{\baselinestretch}%
{#1}\small\normalsize} \spacingset{1}


\if1\blind
{
  \title{\bf Title}
  \author{Author 1\thanks{
    The authors gratefully acknowledge \textit{please remember to list all relevant funding sources in the unblinded version}}\hspace{.2cm}\\
    Department of YYY, University of XXX\\
    and \\
    Author 2 \\
    Department of ZZZ, University of WWW}
  \maketitle
} \fi

\if0\blind
{
  \title{\bf A Kernel-Based Neural Network Test for High-dimensional Sequencing Data Analysis}
  \author{Tingting Hou, Chang Jiang and Qing Lu}
  \maketitle
} \fi

\bigskip
\begin{abstract}
The recent development of artificial intelligence (AI) technology, especially the advance of deep neural network (DNN) technology, has revolutionized many fields. While DNN plays a central role in modern AI technology, it has been rarely used in sequencing data analysis due to challenges brought by high-dimensional sequencing data (e.g., overfitting). Moreover, due to the complexity of neural networks and their unknown limiting distributions, building association tests on neural networks for genetic association analysis remains a great challenge. To address these challenges and fill the important gap of using AI in high-dimensional sequencing data analysis, we introduce a new kernel-based neural network (KNN) test for complex association analysis of sequencing data. The test is built on our previously developed KNN framework, which uses random effects to model the overall effects of high-dimensional genetic data and adopts kernel-based neural network structures to model complex genotype-phenotype relationships. Based on KNN, a Wald-type test is then introduced to evaluate the joint association of high-dimensional genetic data with a disease phenotype of interest, considering non-linear and non-additive effects (e.g., interaction effects). Through simulations, we demonstrated that our proposed method attained higher power compared to the sequence kernel association test (SKAT), especially in the presence of non-linear and interaction effects. Finally, we apply the methods to the whole genome sequencing (WGS) dataset from the Alzheimer's Disease Neuroimaging Initiative (ADNI) study, investigating new genes associated with the hippocampal volume change over time. 
\end{abstract}

\noindent%
\vfill

\newpage
\spacingset{1.45} 
\section{Introduction}
\label{sec:intro}
Utilizing human genome findings and other reliable risk predictors for early disease detection is key in the evolution towards precision medicine. Rapid advancements in high-throughput sequencing technologies, along with consistently decreasing costs, have enabled the gathering of extensive genetic data. This opens the door for a thorough examination of a wide array of human genomic variations in the risk prediction of complex diseases. However, this high-dimensional sequencing data brings unique computational and statistical challenges \cite{fan2014challenges}. One of the most significant challenges is the "curse of dimensionality". This phenomenon occurs when, as the number of dimensions (or features) in a dataset increases, the volume of the space expands so quickly that the data available for analysis becomes increasingly sparse\cite{koutroumbas2008pattern}. The curse of dimensionality negatively impacts the generalizability of algorithms, especially when they are trained and evaluated on insufficient sample sizes for complex constructs like human health, leading to highly variable and often misleading estimates of true model performance. This variability hampers the ability to accurately predict model performance on new data, potentially resulting in catastrophic failures upon deployment if the model's effectiveness is overestimated \cite{berisha2021digital}. 

Deep learning methods have shown significant promise in analyzing high-dimensional sequencing data, often outperforming traditional machine learning methods in several aspects \cite{angermueller2016deep}. Deep neural networks facilitate bypassing manual feature extraction by learning from data and capturing nonlinear dependencies in the sequence, interaction effects, and spanning broader sequence context across multiple genomic scales via complex models that derive data representation through multiple layers of abstraction \cite{angermueller2016deep}. The effectiveness of deep learning hinges on the development of specialized neural network architectures capable of reflecting crucial data properties, such as spatial locality (convolutional neural networks – CNNs), sequential characteristics (recurrent neural networks – RNNs), context dependence (Transformers), and data distribution (autoencoders – AEs). These deep learning models have revolutionized domains such as speech recognition, visual object recognition, and object detection while also playing a pivotal role in addressing significant challenges in computational biology. The application of deep learning in various computational biology areas, like functional biology, is expanding, while its adoption in fields like phylogenetics is still in its early stages \cite{sapoval2022current}. Nevertheless, neural networks also exhibit certain limitations in genomics, with their inherent "black box" nature posing interpretability and comprehension challenges for underlying biological mechanisms \cite{chakraborty2017interpretability}. Therefore, building significance tests on neural networks to examine complex genotype-phenotype relationships remains a great challenge.

In our previous work, we introduced a kernel-based neural networks method that combines the features of linear mixed models (LMM) and classical neural networks, offering a unique structure for analyzing complex genotype-phenotype relationships in high-dimensional sequencing data \cite{shen2021kernelbased}. This approach has demonstrated improved prediction accuracy when nonlinear activation functions are applied. Inference-based methods in genomics are essential for advancing knowledge by uncovering underlying biological mechanisms and establishing causative relationships, which in turn facilitate the identification of key genes, pathways, and novel gene interactions, ultimately paving the way for targeted therapeutic strategies and the discovery of potential drug targets. 

In this paper, we build upon the kernel-based neural network framework to propose a Wald-type test that assesses the joint association of a set of genetic variants with a disease phenotype, accounting for non-linear and non-additive effects. Furthermore, we present two tests that evaluate linear genetic effects and non-linear/non-additive genetic effects (e.g., interaction effects), respectively.  Comprehensive simulations were performed to evaluate the improvement of the newly proposed method in accuracy and power over the sequence kernel association test(SKAT) \cite{wu2011rare}. The new testing method was then used to analyze the change in hippocampus volume over a period of 12 months in Alzheimer’s Disease Neuroimaging Initiative (ADNI)\cite{mueller2005alzheimer}.

\section{Methodologies}
\label{sec:meth}
\subsection{Kernel-Based Neural Network}
\begin{figure}[htbp]
\centering
\includegraphics[width=\textwidth]{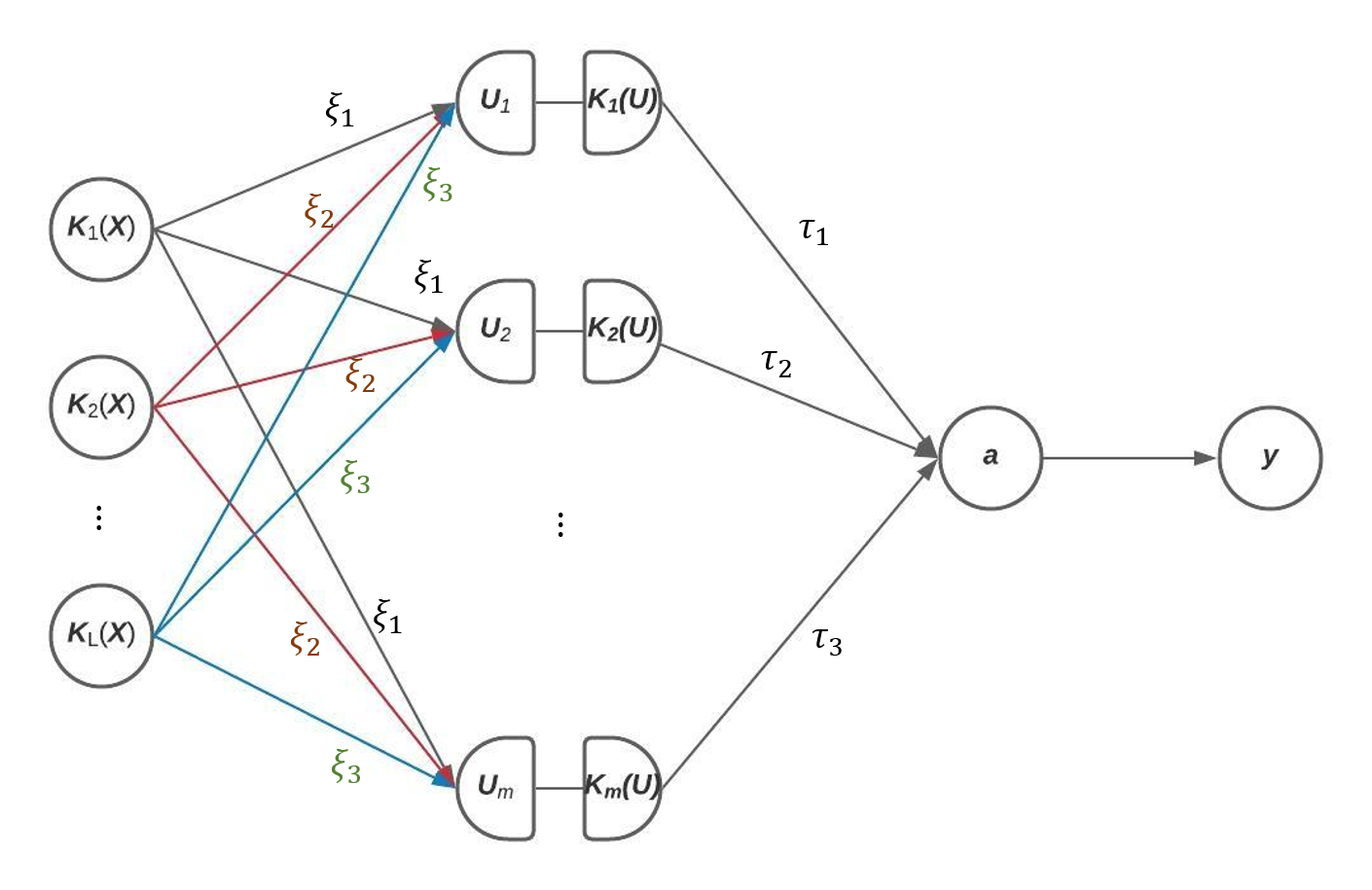}
\caption{An illustration of the hierarchical structure of the kernel neural network model.}\label{fig:KNNfig}
\end{figure}

The kernel-based neural network is a promising method that combines the advantages of kernel methods and mixed linear model, trying to capture the underlying information and improving the accuracy of genetic risk prediction. Figure \ref{fig:KNNfig} shows a basic hierarchical structure of the kernel-based neural network. The phenotype  $\bm{y}$ is modeled as random effect model given some hidden variables $\bm{u_1}, \dots, \bm{u_m}$,
\begin{align}
        \bm{y|Z,a} &\sim  \mathcal{N}_n \left(\bm{Z}\bm{\beta}+\bm{a},\phi \bm{I}_n \right)   \\
        \bm{a}|\bm{u_1}, \dots, \bm{u_m} &\sim  \mathcal{N}_n\left(\bm{0},\sum_{j=1}^J \tau_j \bm{K}_j(\bm{U})\right) 
\end{align}
where $n$ is the sample size; $m$ is the number of hidden units in the network; $\bm{Z}$ is design matrix, $\bm{\beta}$ is the vector of fix effects; the covariance matrix of random effect $\bm{a}$ is a positive combination of some latent kernel matrices $\bm{K}_j(\bm{U})= f_j \left[ \frac{1}{m} UU^T\right]$ , where $\bm{U} = [\bm{u_1}, \dots, \bm{u_m}] \in \mathbb{R} ^{n \times m}$. The latent variables $\bm{u_i}$ is constructed from
\begin{align}
    \bm{u_1}, \dots, \bm{u_m} \sim  \mathcal{N}_n\left(\bm{0},\sum_{l=1}^L \xi_l \bm{K}_l(\bm{X})\right) 
\end{align}
where $\bm{K}_l(\bm{X})$, l = 1, . . . , L are kernel matrices constructed based on the genetic variables.  Then the marginal mean and variance of $y$ can be written  as
\begin{align}
\mbb{E}[\mbf{y}] & =\mbb{E}\left(\mbb{E}[\mbf{y}|\mbf{u}_1,\ldots,\mbf{u}_m]\right)=\mbf{0}.\\
\mrm{Var}[\mbf{y}] & =\mbb{E}\left[\mrm{Var}\left(\mbf{y}|\mbf{u}_1,\ldots,\mbf{u}_m\right)\right]+\mrm{Var}\left[\mbb{E}\left(\mbf{y}|\mbf{u}_1,\ldots,\mbf{u}_m\right)\right] \nonumber \\
	& =\sum_{j=1}^J\tau_j\mbb{E}[\mbf{K}_j(\mbf{U})]+\phi\mbf{I}_n\\
	& \simeq\tau f[\mbf{\Sigma}]+\phi\mbf{I}_n,
\end{align}
where $\mbf{\Sigma}=\sum_{l=1}^L\xi_l\mbf{K}_l(\mbf{X})$ (Xiaoxi). If $f(x) = (1+x)^2$, which corresponds to the output polynomial kernel, then $f[\mbf{\Sigma}]=(\mbf{J}+\mbf{\Sigma})^{\owedge2}$, where the symbol $\owedge2$ means the element-wise square. Without loss of generality, in this paper, we consider the case with $L=1$ and $\mbf{K}_1(\mbf{X})=p^{-1}\mbf{XX}^T$. Then marginal variance of y can be  written as 
\begin{align}
\mrm{Var}[\mbf{y}] & \simeq\tau f[\mbf{\Sigma}]+\phi\mbf{I}_n \\
	& =\tau\mbf{J}+2\tau\xi_1\frac{1}{p}\mbf{XX}^T+\tau\xi_1^2\frac{1}{p^2}(\mbf{XX}^T)^{\owedge2}+\phi\mbf{I}_n \\
	& =\theta_1 \mbf{J}+\theta_2 \frac{1}{p}\mbf{XX}^T+\theta_3 \frac{1}{p^2}(\mbf{XX}^T)^{\owedge2}+\theta_4 \mbf{I}_n, \label{eq:varKNN}
\end{align}
where the parameters $\theta_0,\ldots,\theta_4$ can be estimated via the Minimum Norm Quadratic Unbiased Estimator (MINQUE) \cite{rao1970estimation, rao1971estimation,rao1972estimation}.
\clearpage
\subsection{Kernel-Based Neural Network Association Test}
The MINQUE  requires only a modest distribution assumption and is considered consistent for any choice of initial values, but the limiting distribution of the estimator may depend on the initial values through the covariance matrix \cite{brown1976asymptotic}. Let $\hat{\mbf{\sigma}}_M$ be the MINQUE of variance components, then $ N^{\frac{1}{2}}(\hat{\mbf{\sigma}} -\mbf{\sigma})$ has a limiting normal distribution if $\mbf{Y}$ is normally distributed,
\begin{align}
    N^{\frac{1}{2}}(\hat{\mbf{\sigma}} -\mbf{\sigma}) \sim N\left(\mbf{0}, \mbf{M}(\mbf{H})\mbf{\Psi} \mbf{M}(\mbf{H})'\right),
\end{align}
where N is the number of independent and identically vector observations on $\mbf{Y}$, which in our case is 1; $\mbf{H} = \sum_{l=1}^L a_i \mbf{V}_i$, where $a_i$ is the initial value for the variance matrix $\mbf{V}_i$; $\mbf{M}(\mbf{H}) = (\mbf{L}'\mbf{\Phi}^{-1}(\mbf{H})\mbf{L})^{-1}\mbf{L}' \mbf{\Phi}^{-1}(\mbf{H})$, where $\mbf{\Phi}(\mbf{H}) = [\phi_{ij,kl}]$, $\phi_{ij,kl} = h_{il}h_{jk}+h_{ik}h_{jl}$, and $\mbf{L}$ is the matrix with $i$th column equal to the vector form of $V_i$; furthermore, $\mbf{\Psi} = \mbf{\Phi}(\mbf{\Sigma})$, where $\mbf{\Sigma} = \sum_{l=1}^L \mbf{V}_i \sigma^2_i$.

Although the MINQUE method is computationally efficient, the calculation of the covariance matrix of the limiting distribution gives rise to heavy computational burdens. Fortunately,  the Iterated-MINQUE, is obtained by repeating the MINQUE estimation procedure using previously estimated values as initial values until converged, providing a special form of limiting distribution that is functionally independent of the choice of $\mbf{H}$. With the Iterated-MINQUE, the initial matrix $\mbf H$ is converge to the estimated variance-covariance matrix $\hat{\mbf \Sigma} = \sum_{l=1}^L \mbf{V}_i \hat{\sigma}^2_i$, and furthermore, $\hat{\mbf \Sigma} = \sum_{l=1}^L \mbf{V}_i \hat{\sigma}^2_i \longrightarrow_p \mbf \Sigma$, the variance-covariance matrix  of  $N^{\frac{1}{2}}(\hat{\mbf{\sigma}} -\mbf{\sigma})$ would  be reduced to  
\begin{align}
    \mbf{M}(\mbf{H})\mbf{\Psi} \mbf{M}(\mbf{H})' &= \mbf{M}(\hat{\mbf{\Sigma}})\mbf{\Psi} \mbf{M}(\hat{\mbf{\Sigma}})' \\
    &= \left(\mbf{L}'\mbf{\Phi}^{-1}(\hat{\mbf{\Sigma}})\mbf{L} \right)^{-1}\mbf{L}' \mbf{\Phi}^{-1}(\hat{\mbf{\Sigma}}) \mbf{\Phi}(\mbf{\Sigma}) \mbf{\Phi}^{-1}(\hat{\mbf{\Sigma}}) \mbf{L} \left(\mbf{L}'\mbf{\Phi}^{-1}(\hat{\mbf{\Sigma}})\mbf{L}\right)^{-1} \\
    &\longrightarrow (\mbf{L}'\mbf{\Phi}^{-1}(\mbf{\Sigma})\mbf{L})^{-1}
\end{align}
Furthermore, Brown \cite{brown1976asymptotic} showed that  $(\mbf{L}'\mbf{\Phi}^{-1}(\mbf{\Sigma})\mbf{L})^{-1}$ could be written $[\frac{1}{2} tr \mbf{\Sigma}^{-1} \mbf{V}_i \mbf{\Sigma}^{-1} \mbf{V}_j ]^{-1}$ if $\mbf{Y}$ is normally distributed.

Therefore, the limiting distribution of $\hat{\mbf{\sigma}}$ could be written as 
\begin{align}
    N^{\frac{1}{2}}(\hat{\mbf{\sigma}} -\mbf{\sigma}) \sim N\left(\mbf{0}, [\frac{1}{2} tr \mbf{\Sigma}^{-1} \mbf{V}_i \mbf{\Sigma}^{-1} \mbf{V}_j ]^{-1}\right), \label{eq:varMNIQUE}
\end{align}
With the equation \ref{eq:varMNIQUE}, the limiting distribution of the estimated $\hat{\theta}$ from the kernel-based neural network framework could be written as
\begin{align}
    (\hat{\mbf{\theta}} -\mbf{\theta}) \sim N\left(\mbf{0}, [\mbf{L}'\mbf{\Phi}^{-1}(\mbf{\Sigma})\mbf{L}]^{-1}\right)
\end{align}
With the limiting distribution \ref{eq:varMNIQUE},  the marginal distributions of the estimated $\mbb \theta$ could be written  
According to the limiting distribution,  the marginal distributions of $\theta_2$ and $\theta_3$ in equation (\ref{eq:varKNN}) could be estimated by replacing $\mbf{\Sigma}$ with $\hat{\mbf{\Sigma}}$. Therefore, the limiting distribution 

Therefore, we can build two z-test statistics, $Z_1$ and $Z_2$ for the parameters $\theta_2$ and $\theta_3$ respectively,
\begin{align}
    z_1=\frac{\hat{\theta_2}}{\sqrt{\hat{\sigma}_2^2}}, z_2=\frac{\hat{\theta_3}}{\sqrt{\hat{\sigma}^2_3}}
\end{align}
where $\hat{\sigma}_2^2$ and $\hat{\sigma}_3^2$ are the estimated variance of $\hat{\theta_2}$ and $\hat{\theta_3}$. Also, a chi-square test with 2 df for the total genetic effects
\begin{align}
      \chi^2_2= \hat{\mbf \theta}^T \mbf V_n(\hat{\mbf \theta})^{-1}  \hat{\mbf \theta}
\end{align}
where $\mbf V_n(\hat{\mbf \theta})$ is the estimated covariance of $\mbf \theta =(\hat{\theta}_2,\hat{\theta}_3)'$.. Furthermore, since the parameters we are interested in are positive variance components, one-sided tests are applied to the null hypothesis $H_0$: $\theta_2=0, \theta_3=0$, as well as the Chi-square test using Follmann's method\cite{follmann1996simple}, in which null hypothesis will be rejected at level $\alpha$ as long as $\chi^2$ exceeds its $2\alpha$ critical value and $\hat{\theta}_2^*+\hat{\theta}_3^*>0$, where $\hat{\theta}_2^*$ and $\hat{\theta}_3^*$ are the new value of $\hat{\theta}_2$ and $\hat{\theta}_3$ after standardization.

\section{Simulation}
In this section,  we conducted some simulations to compare the performance of the proposed method based on the kernel-based neural network to the performance of the sequence kernel association test (SKAT) in the linear mixed model. All the simulations are based on 1000 individuals with 1000 Monte Carlo iterations. Simulations were conducted utilizing genuine genotype data from the UKB. The methodologies for quality control (QC) and a comprehensive description of the data are elaborated in \cite{bycroft2018uk}. Post-QC, 320,021 samples had been genotyped via the UK Biobank Axiom array and the UK BiLEVE Axiom array, culminating in 516,429 retained autosomal SNPs. For simplification, a subset of 61,463 samples was randomly chosen, excluding chromosome 6 due to the intricate architecture of the major histocompatibility complex (MHC) region. Consequently, the simulations were performed using the resultant 61,463 samples and 341,545 SNPs.

In our simulation study, we examine the performances of both methods under the situation of nonlinear random effects and non-additive effects. Specifically, we used the following model to simulate the response:
\begin{align}
    \mbf{y}=f(\mbf{a}) + \mbf{\epsilon}, ,\quad\mbf{a}\sim\mcal{N}_n\left(\mbf{0},\sigma_g^2 \frac{1}{p_{\lambda}}\Tilde{\mbf{G}}  \Tilde{\mbf{G}}^T\right),
\end{align}
where $\Tilde{\mbf{G}}$ is a $n\times p_\lambda$  the casual genotype matrix which is a small proportion $\lambda$, e.g. 20\%, of $n\times p$ SNPs matrix $\mbf{G}$; and $\mbf{\epsilon}\sim\mrm{ i.i.d. }\mcal{N}_n(\mbf{0},\sigma^2_0\mbf{I}_n)$. The trait was simulated under the null model or one of four types of functions $f$: e linear ($f(x)=x$), quadratic ($f(x)=x^2$), hyperbolic cosine ($f(x)=cosh(x)$) and ricker curve ($f(a)=\beta*r(a^2)exp(-r(a^2))$, where $r$ is the soft rectifier, $r(a)=log(1+e^a)$), respectively, to assess the type I error or the statistical power. 

Furthermore,  we explore the performances of both methods under non-additive effects. In simulation studies, we mainly focus on the interaction effect and generate the response using the following model:
$$
\mbf{y}=f(\Tilde{\mbf{G}})+\mbf{\epsilon},
$$
where $\Tilde{\mbf{G}}=[\mbf{g}_1,\ldots,\mbf{g}_{ p_\lambda}]\in\mbb{R}^{n\times  p_\lambda}$ is the SNP data and $\mbf{\epsilon}\sim\mcal{N}_n(\mbf{0},\mbf{I}_n)$. When applying both methods, the mean is adjusted so that the response has a marginal mean of 0. In the simulation, we considered two types of interaction models, the multiplicative interaction and the threshold interaction,  with the following functions.

The multiplicative interaction
\begin{align}
    f(\Tilde{\mbf{G}})=\sum_{1\leq j_1<j_2\leq p_\lambda }\mbf{g}_{i_{j_1}}\odot\mbf{g}_{i_{j_2}}
\end{align}
The threshold interaction
\begin{align}
f(\Tilde{\mbf{G}})= \begin{cases}
    \sum_{1\leq j_1<j_2\leq p_\lambda }\mbf{g}_{i_{j_1}}\odot\mbf{g}_{i_{j_2}}, &\quad \text{if}\quad \sum_{1\leq j_1<j_2\leq p_\lambda }\mbf{g}_{i_{j_1}}\odot\mbf{g}_{i_{j_2}} >0 \\
    0,  &\quad \text{otherwise.}
\end{cases}
\end{align}

where $\odot$ stands for the Hadamard product.

\subsection{Simulation I}
In Simulation I, we aimed to assess the performance of the two methods with varying SNP sizes. The phenotype of 1000 individuals was generated based on 500, 2,000, and 4,000 SNPs randomly selected from the UKB under the null model or one of the models, nonlinear models or non-additive model.  The proportion of causal SNPs was fixed at 20\%.  Table \ref{tab:simIset} presents the parameter settings for all scenarios in Simulation I, where $\sigma_g^2$ represents the genetic effect, $\sigma_0^2$ denotes the noise, and $\beta$ is a parameter specific to the Ricker curve model. Each simulation scenario was replicated 1,000 times. 

\begin{table}[htb]
\centering
\caption{The scenario settings in simulation I}
\label{tab:simIset}
\resizebox{\columnwidth}{!}{%
\begin{tabular}{ccccccccccl}
\cline{1-10}
Sample   Size & \multicolumn{3}{c}{500} & \multicolumn{3}{c}{2000} & \multicolumn{3}{c}{4000} &  \\ \cline{2-10}
 & $\sigma_g^2$ & $\sigma^2_0$ & $\beta$ & $\sigma_g^2$ & $\sigma^2_0$ & $\beta$ & $\sigma_g^2$ & $\sigma^2_0$ & $\beta$ &  \\ \cline{1-10}
Linear & 0.50 & 2.00 & NA & 0.90 & 2.00 & NA & 1.20 & 2.00 & NA &  \\
Quadratic & 1.50 & 2.00 & NA & 5.00 & 2.00 & NA & 7.00 & 2.00 & NA &  \\
Hyperbolic & 2.00 & 2.00 & NA & 3.00 & 2.00 & NA & 3.00 & 2.00 & NA &  \\
Rickercurve & 0.50 & 2.00 & 30.00 & 1.00 & 70.00 & NA & 1.00 & 2.00 & 150.00 &  \\
Multiplicative   interaction (2- way ) & 8.00 & 2.00 & NA & 10.00 & 2.00 & NA & 20.00 & 2.00 & NA &  \\
Multiplicative   interaction (3- way ) & 10.00 & 2.00 & NA & 20.00 & 2.00 & NA & 20.00 & 2.00 & NA &  \\
Threshold   interaction (2-way) & 8.00 & 2.00 & NA & 10.00 & 2.00 & NA & 50.00 & 2.00 & NA &  \\
Threshold   interaction (3-way) & 5.00 & 2.00 & NA & 20.00 & 2.00 & NA & 20.00 & 2.00 & NA &  \\ \cline{1-10}
\end{tabular}%
}
\end{table}

Table \ref{tab:TypeIerror} displays the type I error rates estimated for our proposed method at a significance level of $\alpha = 0.05$ across diverse scenarios of null models for continuous phenotypes. The results indicate that our proposed method is generally protected against type I errors, though for smaller sample sizes, the SKAT method may be slightly conservative. All the Type I error rates were found to be close to the nominal levels, with rates of 0.036, 0.042, and 0.044 for the overall effect, the linear effect, and the non-linear effect when the SNP size was 500, and rates of 0.054, 0.046, and 0.046 for the overall effect, the linear effect, and the non-linear effect respectively, with a 2000 SNP size. However, the SKAT method yielded severely conservative Type I error rates in high-dimensional data analysis. For instance, the type I error rate of the SKAT was 0.014 for a SNP size of 2000 and worsened with the SNP size increasing, with a type I error rate of 0.003 for a SNP size of 4000. These findings demonstrate that MINQUE testing is a valid method and has good power in high-dimensional data analysis compared to existing methods.

\begin{figure}[htb]
  \centering
 \includegraphics[width=\linewidth]{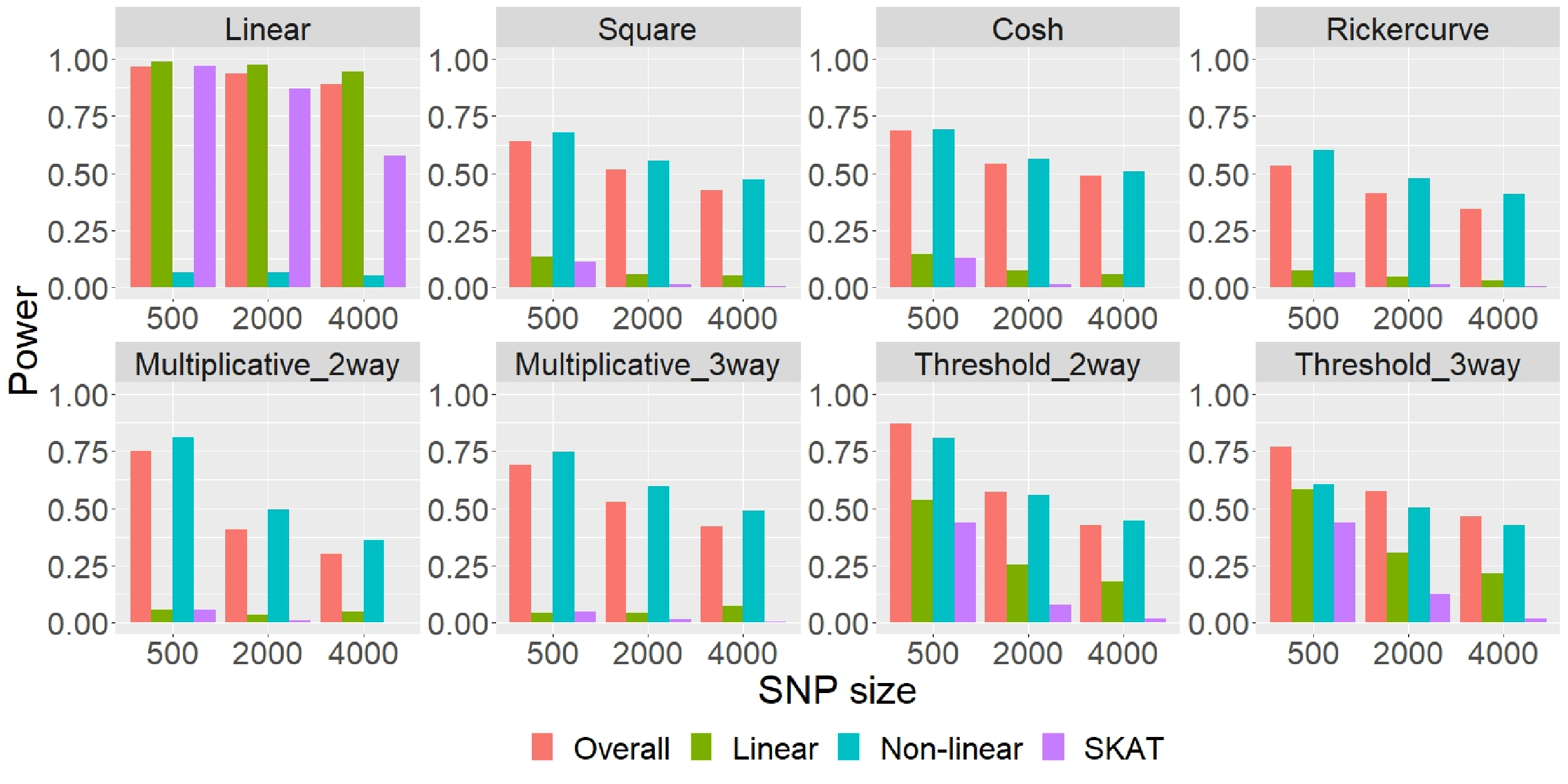}
  \caption{The power comparison between the kernel-based neural network Testing and SKAT with varying SNP size}
  \label{fig:SNP_size}
\end{figure}

Furthermore, We summarize the statistical power of our proposed testing and SKAT via barplot in Figure \ref{fig:SNP_size}.  As we observe from the bar plot, under the linear scenarios, the overall genetic powers of our proposed method are comparable to the SKAT method when the SNP size is smaller than the sample size. In linear scenarios with 500 SNPs, the overall effect testing power for the proposed method and SKAT at a 0.05 significance level were 96.8\% and 96.9\%, respectively. As the number of variants increases, our method's performance outperformed the SAKT, particularly with 4,000 SNPs, where the overall effect power and linear effect power of our method reached 89.0\% and 94.5\%, respectively, compared to SKAT's 57.6\%.

In non-linear scenarios, the SKAT method struggles to detect signals with a product kernel, while our method can still differentiate linear and non-linear effects. For example, using the square function and 500 SNPs, the overall effect, linear effect, and non-linear effect powers were 64.2\%, 13.5\%, and 67.8\%, respectively, while SKAT's power was only 11.3\%. Similar trends were observed in other non-linear/non-additive scenarios.

\begin{table}[htbp]
\centering
\caption{The Type I Error comparison between the kernel-based neural network  Testing and SKAT under significant level 0.05}
\label{tab:TypeIerror}
\begin{tabularx}{\textwidth}{ X  X  X  X X }
\hline
\textbf{SNP} & \textbf{Total} & \textbf{$\theta_2$} & \textbf{$\theta_3$} & \textbf{SKAT} \\ \hline
500          & 0.036          & 0.042        & 0.044        & 0.042         \\
2000         & 0.054          & 0.046        & 0.046        & 0.014         \\
4000         & 0.050           & 0.047        & 0.056        & 0.003         \\ \hline
\end{tabularx}
\end{table}

\clearpage
\subsection{Simulation II}
\begin{figure}[htb]
  \centering
  \includegraphics[width=\linewidth]{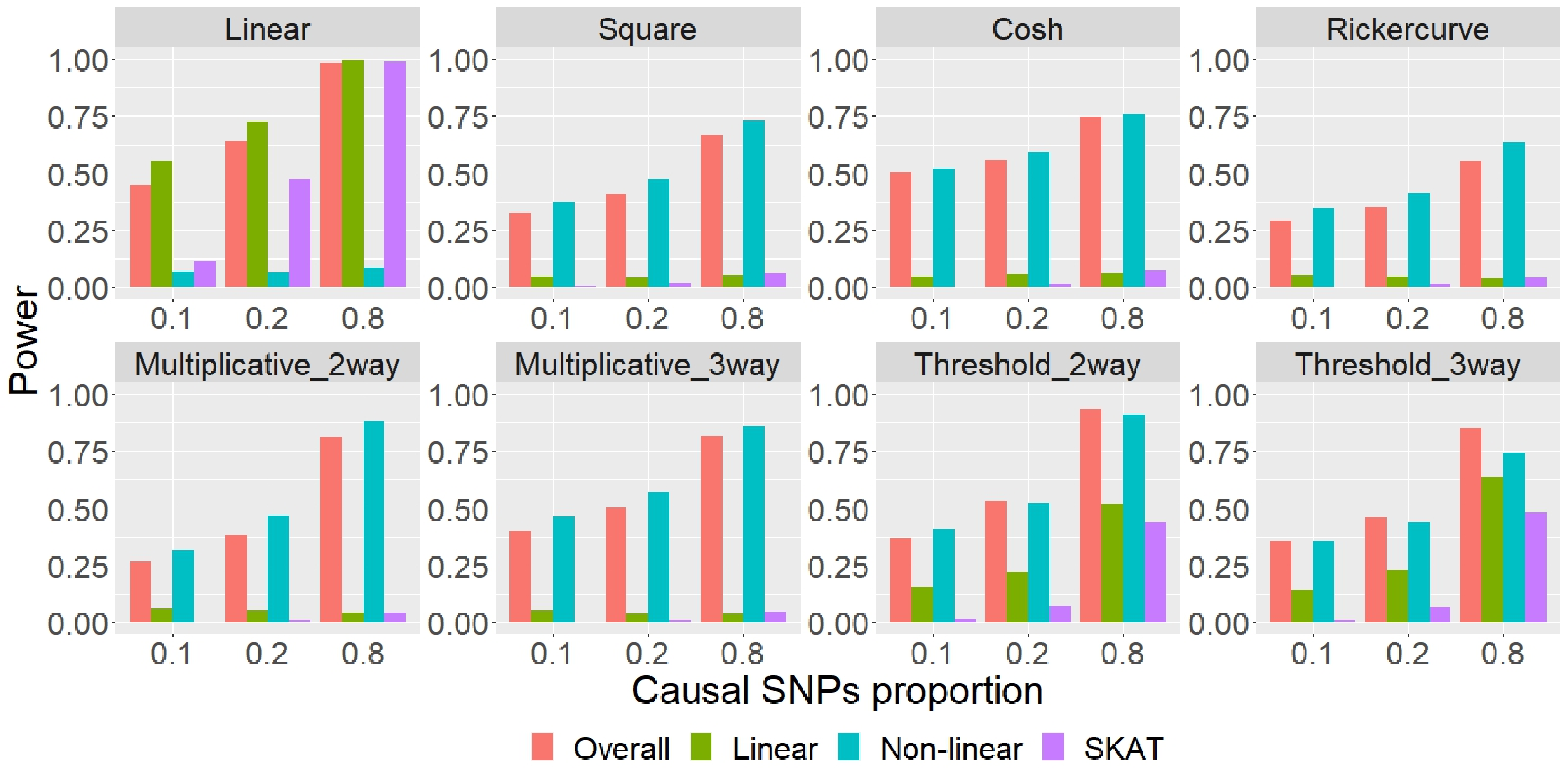}
  \caption{The power comparison between the kernel-based neural network Testing and SKAT with varying proportions of  causal SNP }
  \label{fig:SNP_ratio}
\end{figure}

In this simulation, we investigated the performance of two methods under different proportions of causal SNPs. 1,000 samples, each with 500 SNPs randomly obtained from the UKB dataset, were generated with the parameter setting in \ref{tab:simIset} under various simulation models, including linear/non-linear models and non-additive models. For this simulation, the causal SNP proportions were set to 10\%, 20\%, and 80\%.  Each simulation scenario was replicated 1,000 times. The results are shown in Figure \ref{fig:SNP_ratio}.

Figure \ref{fig:SNP_ratio} demonstrates the relationship between statistical power and the proportion of causal SNPs across various genetic models. As the proportion of causal SNPs increases, the statistical power for all methods generally improves.Notably, our method consistently showcases superior statistical power across varying proportions of causal SNPs.  Within the linear scenarios, SKAT's statistical power undergoes a steep decline as the causal SNP proportion transitions from 80\% to 10\%. In Contrast, our proposed methodology exhibits a more gradual decrement in power. Specifically, where SKAT's power diminishes from 98.7\% to 11.7\%, our method's overall genetic effect testing evidences a reduction to  44.5\%. Furthermore, for the linear effect testing, our method's power reduced from 99.5\% to 55.4\%, which also shows the robustness of our proposed method against the different proportions of causal SNPs.

For non-linear and non-additive genetic models, our proposed methodology persistently surpasses SKAT. Figure \ref{fig:SNP_ratio} shows that SKAT encounters challenges in exploring these complex genetic effects, whereas our method manifests consistent robustness. The disparity in statistical power between the two methods becomes increasingly significant in these scenarios, underscoring the superior efficacy of our proposed approach in accommodating complex genetic effects.

\clearpage
\subsection{Simulation III}
In this simulation, we consider different types of weight functions and explore the performance in terms of statistical powers between our proposed method and SKAT. Both MINQUE testing and SKAT can employ pre-specified weights to enhance the power of association tests, and their performance may depend on the degree to which the weights reflect the genetic variants' relative contributions to the disease. In the absence of prior knowledge of the underlying disease model, weights are often pre-specified as a function of MAFs. In this simulation, we selected four weight functions, unweighted(UW), beta distribution (BETA), weighted sum statistics(WSS)  and logarithm of MAFs, and evaluated their impact on the methods' performance. 

\begin{figure}[htb]
  \centering
  \includegraphics[width=\linewidth]{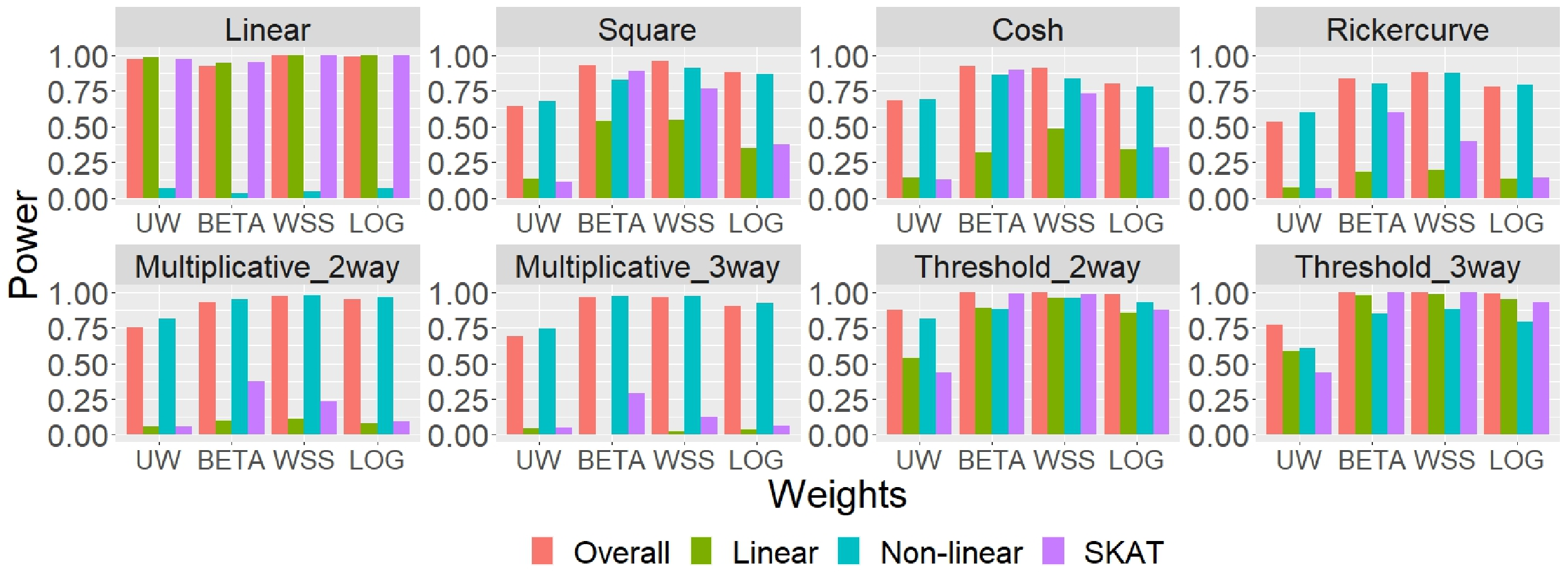}
  \caption{The power comparison between the kernel-based neural network Testing and SKAT with rare variants using varying different weights}
  \label{fig:SNP_rare}
\end{figure}

To compare the two methods' power, we randomly selected 500 SNPs as causal variants with a 20\% causal proportion from the UKB dataset and simulated their phenotypes according to the specified models,  including linear/non-linear models and non-additive models. In this simulation, we selected four weight functions, unweighted(UW), beta distribution (BETA), weighted sum statistics(WSS)  and logarithm of MAFs, and evaluated their impact on the methods' performance.  The parameter settings for each model are provided in Table \ref{tab:simIset}.  Figure \ref{fig:SNP_rare} summarizes simulation results for various weight functions using UKB data. 

Figure \ref{fig:SNP_rare} showcases the comparative advantages of our proposed method over SKAT across various genetic models and weight functions. In a majority of genetic models, such as Linear, Square, Cosh, and Rickercurve, our method consistently demonstrates superior or equivalent statistical power, irrespective of the weight function utilized. In the context of the Linear model, our method's efficacy parallels that of SKAT, making the weight function's choice largely irrelevant. Furthermore, in nonlinear or nonadditive models, the choice of weight function notably affects SKAT's power, yet our approach remains steadfast in its robustness. For instance, in the Square model, SKAT's power varies considerably, ranging from 88.7\% using the BETA function to just 11.3\% when unweighted. Conversely, our proposed method displays consistent power across various weight functions, evidenced by a range from 95.4\% with the BETA function down to 64.2\% when unweighted. These observations are consistent with trends seen in other simulation scenarios. Such unwavering performance underscores the efficacy and resilience of our method, particularly in scenarios typified by rare genetic variants, regardless of weight function variations.

\clearpage
\section{Real Data Application}

The Alzheimer's Disease Neuroimaging Initiative (ADNI) is a collaboration between the public and private sectors that was established in 2003 to evaluate the effectiveness of various biomarkers in tracking the progression of Mild Cognitive Impairment (MCI) and Alzheimer's Disease (AD). The biomarkers include serial MRI and PET scans, biological markers, as well as clinical evaluations and neuropsychological assessments. The ADNI cohort has been continuously enrolling participants from over 60 sites across the United States and Canada, and is divided into four phases (ADNI-1, ADNI-2, ADNIGO, and ADNI-3). The participants are categorized into three groups, including individuals with no memory loss (CN), those who meet the criteria for probable AD outlined by the National Institute of Neurological and Communicative Disorders and Stroke-Alzheimer’s Disease and Related Disorders Association (NINCDS-ADRDA), and individuals with Mild Cognitive Impairment (MCI). The diagnostic criteria used by ADNI are detailed in \cite{petersen2010alzheimer}.

In this study, we examined the change in hippocampus volume over a period of 12 months in two ADNI study groups using our method to identify genes associated with variations in hippocampal volume. We analyzed the genome sequencing data of 808 participants at the baseline of the ADNI1 and ADNI2 studies. We excluded SNPs with low calling rates ($<$0.9), low minor allele frequencies (MAF) ($<$0.01), or those that did not pass the Hardy Weinberg exact test ($p$-value$<$1e-6), as well as non-European American samples. The remaining data was uploaded to the University of Michigan server for posterior likelihood imputation (\url{https://imputationserver.sph.umich.edu/index.html}). We tested 291,872 SNPs for their association with hippocampus volume expansion and assigned them to each of the 24,219 autosomal genes based on their positions on the human assembly GRCh37 (also known as hg19) in Ensembl (\url{https://useast.ensembl.org/index.html}). To capture regulatory regions and SNPs in linkage disequilibrium, we defined gene boundaries as $\pm$ 5 kb of 5' and 3' UTRs. Normalized product kernel matrices were constructed for each gene with p SNPs for analysis. We also considered age, gender, race, education, APOE4, and the top 10 PCs as covariates.

In this study, we investigated the top 10 significant single nucleotide polymorphisms (SNPs) associated with hippocampus expansion using our proposed testing method. Table 1 presents the results, including Chromosome (Chr), Overall p-value, linear effect p-value ($\theta_2$), non-linear effect p-value ($\theta_3$), P-value from Sequence Kernel Association Test (SKAT), and the corresponding gene (GEN). The highest significance was observed for SNPs located on chromosomes 8 and 15, with p-values of 1.20E-88 and 1.21E-88, respectively. These SNPs were associated with the PVT1 and DDX11L9 genes. Other highly significant SNPs were identified on chromosomes 19, 2, 3, 8, 3, 22, 1, and 15, linked to NFKBID, MRPS5, KCNH8, DEFB104A, TMEM212, POM121L1P, KTI12, and MYEF2 genes.

Upon analyzing the linear ($\theta_2$) and non-linear ($\theta_3$) effect p-values, we observed that all linear effect p-values ($\theta_2$) were not significant, except for the PVT1 gene, whereas the non-linear effect p-values ($\theta_3$) were all extremely significant. These results indicate that the identified genes play a non-linear role in hippocampus expansion. However, when using the SKAT, we found that none of the p-values were significant, suggesting that SKAT was unable to confirm the associations detected by our proposed testing method. This finding demonstrates the superiority of our method in detecting these associations, as it outperformed the traditional SKAT approach. The robust performance of our method highlights its potential for further application in studies aiming to uncover the genetic basis of complex traits and diseases.

\begin{figure}[htb]
  \centering
  \includegraphics[width=\linewidth]{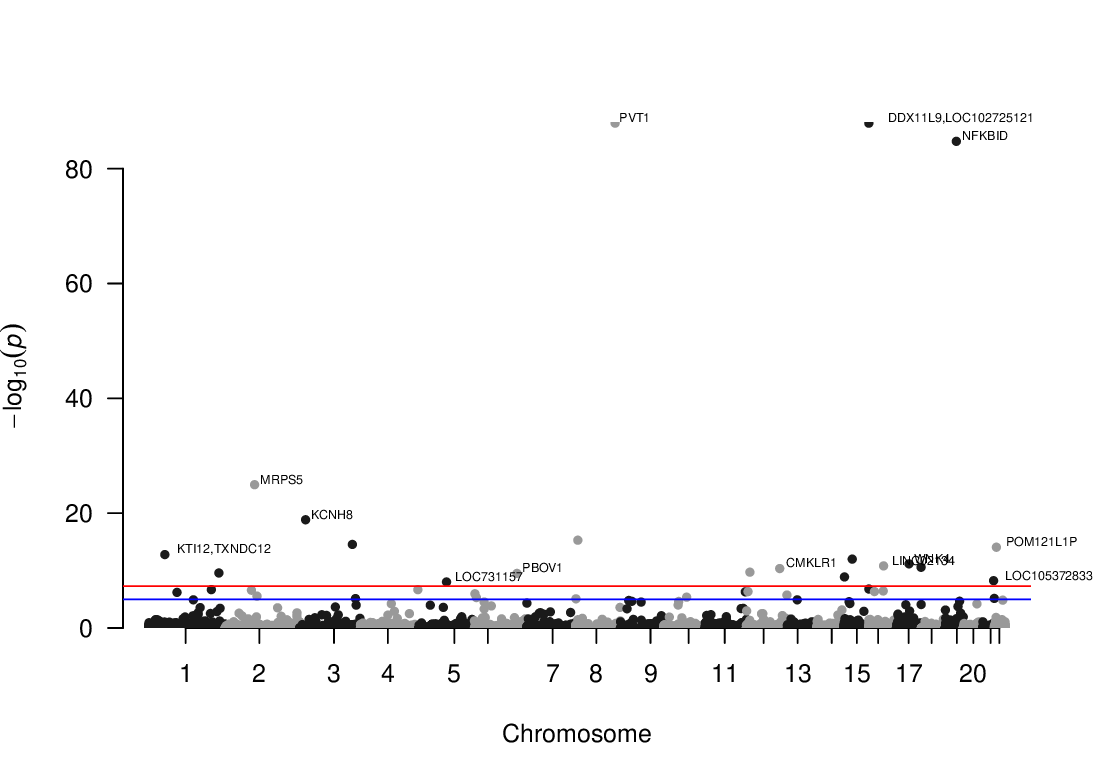}
  \caption{Manhattan plot. The red line represents the genome-wide significance threshold of 5.00E-08, while the blue line corresponds to the suggestive threshold of p = 1.00E-05.}
  \label{fig:Manhattan}
\end{figure}



\begin{table}[!ht]
    \centering
    \caption{Top 10 significant SNP associated with Hippocampus expansion using the kernel-based neural network  Testing }
    \begin{tabular}{|l|l|l|l|l|l|}
    \hline
        \textbf{Chr} & \textbf{Overall} & \textbf{$\theta_2$} & \textbf{$\theta_3$} & \textbf{SKAT} & \textbf{GEN} \\ \hline
        8 & 1.20E-88 & 4.69E-06 & 3.73E-66 & 0.963381765 & PVT1 \\ \hline
        15 &  1.21E-88 & 1 & 8.14E-75 & 0.902781534 & DDX11L9 \\ \hline
        19 &  1.73E-85 & 1 & 3.38E-63 & 0.544132942 & NFKBID \\ \hline
        2 &  1.11E-25 & 0.177086006 & 1.48E-24 & 0.274812148 & MRPS5 \\ \hline
        3 &  1.41E-19 & 1 & 1.50E-16 & 0.707741746 & KCNH8 \\ \hline
        8&	5.11E-16&	1&	2.90E-10&	0.132512299&	DEFB104A\\ \hline
3	&2.82E-15	&0.999999941&	1.40E-14&	0.149287032&	TMEM212\\ \hline
22 &	8.54E-15&	1&	2.50E-10&	0.172189699&	POM121L1P\\ \hline
1	&1.63E-13	&1	&1.99E-09&	0.725531462&	KTI12 \\ \hline
15	&1.01E-12	&1	&1.10E-13&	0.356797849&	MYEF2\\ \hline
    \end{tabular}
    \label{tab:Top5SignificantSNPs}
\end{table}

\clearpage
\section{Discussion}
\label{sec:discussion}
Our study presents a novel association testing method based on the kernel-based neural network framework for evaluating the joint association of multiple genetic variants with a disease phenotype. The Wald-type test we proposed offers a powerful means to evaluate these associations, taking into account non-linear and non-additive effects, which are frequently overlooked in traditional genetic association studies such as the SKAT method. Moreover, we introduced two separate tests specifically designed to assess linear genetic effects and non-linear/non-additive genetic effects (e.g., interaction effects). This innovative approach overcomes the limitations of SKAT methods, which may not sufficiently capture complex genetic interactions or non-linear/non-additive genetic effects. Our kernel-based neural network framework facilitates a more comprehensive understanding of the genetic architecture underpinning disease phenotypes, setting the stage for future research to delve into the complex relationships between genetic variants and diseases.

The simulation studies and the real data application show the advantages of our proposed method. In linear models, the power of our proposed method is comparable to the power of the SKAT under lower-dimension scenarios. However, it surpasses SKAT in high-dimensional situations. This is attributed to the SKAT's conservative tendency, which intensifies as genetic variant numbers grow, a consequence of the second-order bias emerging when replacing $\sigma_0^2$ with $\hat{\sigma}_0^2$ during Davies method application. Conversely, our proposed method effectively controls the type I error in high-dimensional scenarios, highlighting its superiority in handling complex genetic associations. Moreover, our simulations indicate that our method detects subtle genetic signals more effectively than SKAT. Notably, with rare genetic variants, our method consistently outshines SKAT, proving its robustness and efficiency regardless of the chosen weight functions.

In our real data application, we identified numerous genes associated with changes in hippocampal volume over a 12-month period. The top 10 significant genes, listed in Table \ref{tab:Top5SignificantSNPs}, primarily exhibit non-linear effects, indicating that these genes might indirectly influence hippocampal expansion. While the specific pathways through which these genes impact hippocampal growth remain unknown, consistent evidence supports their involvement. Notably, the PVT1 gene, located at the 8q24 chromosomal band, has been extensively studied for its role in various human cancers\cite{guan2007amplification,cui2016long,carramusa2007pvt,liu2015overexpression,traversa2022unravelling}, but recent findings suggest it may also be implicated in neuronal development. Animal studies have revealed that inhibiting PVT1 may decrease neuron loss, suppress astrocyte activation, and increase hippocampal brain-derived neurotrophic factor (BDNF) expression by downregulating the Wnt signaling pathway\cite{zhao2019silencing}. Furthermore, higher BDNF levels have been associated with larger hippocampal volumes, despite age-related decline. Another study demonstrated that PVT1-mediated autophagy might protect hippocampal neurons from synaptic plasticity impairment and apoptosis, thereby ameliorating cognitive impairment in diabetes\cite{li2016autophagy}. These findings collectively suggest that PVT1 could play a crucial role in regulating the expression of other genes through yet-to-be-discovered pathways, ultimately affecting hippocampal neuron development and hippocampal volume.

Despite the identification of other significant genes, the relationship between these genes and hippocampal volume changes remains unclear. However, existing research indicates that these genes may be involved in nervous system activity through specific underlying pathways. For instance, the MRPS5 gene, also known as Mitochondrial Ribosomal Protein S5, encodes a protein vital for the mitochondrial ribosome. Studies have demonstrated that MRPS5 influences neurological stress intolerance and anxiety-related behavioral alterations in mice \cite{akbergenov2018mutant, shcherbakov2021mitochondrial}, and its expression was found to be significantly decreased in the hippocampus of depressed (CUMS) rats \cite{zhang2019itraq}. Additionally, the knockdown of mrps5 has been shown to regulate the expression of bZIP transcription factors ATF5 and ATF4 \cite{zhu2021mitochondrial}. ATF4 is a crucial regulator of the physiological state necessary for neuronal plasticity and memory \cite{pasini2015specific}, while ATF5 may serve a neuroprotective role by decreasing endoplasmic reticulum stress-induced apoptosis in certain hippocampal neuronal fields and primary neuronal culture models \cite{torres2013protective}. Increased ATF5 levels were also observed in apoptosis-resistant epileptogenic foci from patients with temporal lobe epilepsy\cite{torres2013protective}. Another example is KCNH8, a member of the human Elk K+ channel gene family, which plays a critical role in a wide range of physiological processes. Voltage-gated potassium channels, representing the most complex class of voltage-gated ion channels, functionally and structurally, are known to regulate numerous processes such as neurotransmitter release, heart rate, insulin secretion, neuronal excitability, epithelial electrolyte transport, and smooth muscle contraction \cite{zou2003distribution}.

In this study, we introduced an association testing method based on a kernel-based neural network, achieving greater power compared to the sequence kernel association test (SKAT), particularly when confronted with non-linear and interaction effects. it is crucial to recognize certain limitations of our methodology. First, the iterative MINQUE method necessitates significant computational resources, which may impede its application to large datasets, such as the UK Biobank (UKB) dataset. Second, while the MINQUE method does not require the assumption of normality, the limiting distribution still relies on the normality assumption. This dependence may constrain the method's robustness and applicability in situations where the data deviates from normality. Lastly, our method may face limitations under low-dimensional scenarios, where the number of genetic variants is much smaller than the sample size. In such cases, the limiting distribution may not asymptote to the normal distribution, potentially compromising the accuracy of our approach. Consequently, future work should address these limitations to improve the generalizability and effectiveness of the proposed method.

\bibliography{reference}
\end{document}